\documentclass{article}

\usepackage[final]{nips_2016}
\usepackage[utf8]{inputenc} 
\usepackage[T1]{fontenc}    
\usepackage{hyperref}       
\usepackage{url}            
\usepackage{booktabs}       
\usepackage{amsfonts}       
\usepackage{nicefrac}       
\usepackage{microtype}      
\usepackage{graphicx}
\usepackage{amsmath}
\usepackage{multirow}
\usepackage{caption}
\usepackage{algorithm}
\usepackage{algpseudocode}
\usepackage{float}
\usepackage{subfigure}
\usepackage{url}
\newcommand{\tabincell}[2]{\begin{tabular}{@{}#1@{}}#2\end{tabular}}
\title{e-Distance Weighted Support Vector Regression}

\author{
  Yan Wang \\
  Jilin University \\
  Jilin University, Changchun, China \\
  \texttt{wy6868@jlu.edu.cn} \\
  \And
  Ge Ou \\
  Jilin University \\
  Jilin University, Changchun, China \\
  \texttt{ccouge@126.com} \\
  \And
  Wei Pang \\
  University of Aberdeen \\
  University of Aberdeen, Aberdeen, Scotland, U.K. \\
  \texttt{pang.wei@abdn.ac.uk} \\
  \And
  Lan Huang \\
  Jilin University \\
  Jilin University, Changchun, China \\
  \texttt{huanglan@jlu.edu.cn} \\
  \And
  George Macleod Coghill\\
  University of Aberdeen \\
  University of Aberdeen, Aberdeen, Scotland, U.K. \\
  \texttt{g.coghill@abdn.ac.uk} \\
}

\begin{document}
\maketitle
\begin{abstract}
We propose a novel support vector regression approach called e-Distance Weighted Support Vector Regression
(e-DWSVR). e-DWSVR specifically addresses two challenging issues in support vector regression:
first, the process of noisy data; second, how to deal with the situation when the distribution of
boundary data is different from that of the overall data. The proposed e-DWSVR optimizes the minimum
margin and the mean of functional margin simultaneously to tackle these two issues.
In addition, we use both dual coordinate descent (CD) and averaged stochastic gradient descent (ASGD)
strategies to make e-DWSVR scalable to large-scale problems. We report promising results obtained by
e-DWSVR in comparison with existing methods on several benchmark datasets.
\end{abstract}

\section{Introduction}
\label{headings}
\label{gen_inst}
Support Vector Regression (SVR) has recently received much attention due to its competitive performance [1][2] compared with other regression approaches,
including the method of least squares [3], the XCSF algorithm [4], the logistic regression [5], and Neural Networks (NN) [6].
In this paper, we aim to develop a novel SVR approach by considering the recent progress in the support vector (SV) theory and addressing the limitations of existing SVR approaches.

In general, SVR constructs decision functions in high dimensional space for linear regression while the training data are mapped to a higher dimension in kernel Hilbert space.
$\varepsilon$-SVR [7][8] (the symbol $\varepsilon$ is replaced by e in the rest of this paper, and thus e-SVR is used) is the first popular SVR strategy and the other one is v-SVR [9]. e-SVR aims to find a function whose deviation is not more than e, thus forming the so-called e-tube, to fit all training data. In order to find the best fitting surface, e-SVR tries to maximize the minimum
margin containing data points in the e-tube as much as possible, which is similar to SVM. As for v-SVR, it adds a parameter v to the original e-SVR to control the number of support vectors and adjust
the parameter e automatically [10].

However, both e-SVR and v-SVR are susceptible  to the data on the boundary (i.e. the support vectors). In fact, the optimization objective greatly depends on the margin between support vectors,
and this makes the final fitting function heavily rely on the distribution of the support vectors: if the distribution of the whole data within the e-tube is very different from the direction
of the support vectors, the final regression function may not be reliable.

Recent progress in the SV theory suggests that maximizing the minimum margin, that is, the shortest distance from the instances to the separating hyperplane, is not the only optimization goal for
achieving better learning performance. Different from traditional SVMs, Distance-weighted Discrimination (DWD)[11] maximizes the mean of the functional margin of all data [12],
and it uses the distances of all data to define the separating hyperplane,
thus greatly improving the performance. Meanwhile, Large Margin Distribution Machine (LDM) [13][14] is based on the novel theory of minimizing the margin distribution,
and it employed the dual coordinate descent (CD) and the averaged stochastic gradient descent (ASGD) strategies to solve the optimization function.

Considering the above, we propose a novel SVR approach called e-Distance Weighted Support Vector Regression (e-DWSVR).
e-DWSVR optimizes the minimum margin and the mean of functional margin at the same time, and it also uses both CD and ASGD strategies as in LDM to improve its scalability. A comparison of our e-DWSVR
with several other popular regression methods (including e-SVR, NN, linear regression and logistic) indicates that our e-DWSVR fits better the whole data distribution in most cases,
especially for those datasets with strong interference noise.

\section{Background on support vector theory}
Let $S = \left\{ {({x_1},{y_1}),({x_2},{y_2}),...,({x_n},{y_n})} \right\}$  be a training set of $n$ samples, where ${x_i} \in {R^m}$ are the input samples and ${y_i} \in R$ are
the corresponding target values. For regression, the objective function is $f(x) = w \cdot \phi (x) + b$, where $w \in {R^m}$ and $\phi$ is the mapping function induced by a kernel $K$,
i.e., $K({x_i},{x_j}) = \phi ({x_i}) \cdot \phi ({x_j})$, which projects the data to a higher dimensional space.
\subsection{The e-SVR process.}
According to [7][8], the objective function $f(x)$ is represented by the following constrained minimization problem:
\[\begin{array}{l}
\mathop {\min }\limits_{w,{\bf{\xi }},{{\bf{\xi }}^*}} \frac{1}{2}{\left\| w \right\|^2} + C\sum\limits_{i = 1}^n {({\xi _i} + \xi _i^*)} \\
s.t.{\rm{\ }}{y_i} - w \cdot \phi ({x_i}) - b \le \varepsilon  + {\xi _i},{\rm{\  }}w \cdot \phi ({x_i}) + b - {y_i} \le \varepsilon  + \xi _i^*{\rm{,\ \ }}{\xi _i},\xi _i^* \ge 0,{\rm{\  }}i = 1,2,...,n.
\end{array}\]
In the above, $C$ is a parameter that denotes the trade-off between the flatness of $f(x)$ and sums up to which deviations larger than e are tolerated. ${\bf{\xi }} = {[{\xi _1},{\xi _2},...,{\xi _n}]^T}$
and  ${{\bf{\xi }}^{\bf{*}}} = {[\xi _1^*,\xi _2^*,...,\xi _n^*]^T}$ are regarded as the slack variables measuring the distances of the training samples lying outside the e-tube
from the tube itself. After applying the Lagrange multiplier, the minimization problem can be handled as the dual optimization problem:
\[\begin{array}{l}
\mathop {\max }\limits_{\alpha ,{\alpha ^*}}  - \frac{1}{2}{\left( {\alpha  - {\alpha ^*}} \right)^T}K\left( {{x_i},{x_j}} \right)\left( {\alpha  - {\alpha ^*}} \right) - \varepsilon \sum\limits_{i = 1}^n {\left( {{\alpha _i} + \alpha _i^*} \right)}  + \sum\limits_{i = 1}^n {{y_i}\left( {{\alpha _i} - \alpha _i^*} \right)} \\
s.t.{\rm{\  }}\sum\limits_{i = 1}^n {\left( {{\alpha _i} - \alpha _i^*} \right)}  = 0,{\rm{\ \ }}0 \le {\alpha _i},\alpha _i^* \le C,{\rm{\ \ }}i = 1,2,...,n.
\end{array}\]
And the final objective function becomes $f(x) = \sum\limits_{i = 1}^n {({\alpha _i} - \alpha _i^*)} K({x_i},x) + b. $
\subsection{Recent progress in the SV theory}
SVM aims to maximize the minimum margins, which denotes the smallest distances of all instances to the separating
hyperplane [1][2]. The optimization problem is represented as follows:
\[\mathop {\min }\limits_{w,{\bf{\xi }}} \frac{1}{2}{\left\| w \right\|^2} + C\sum\limits_{i = 1}^n {{\xi _i}} {\rm{\ \ \ }}s.t.{\rm{\ }}{y_i}\left( {w \cdot \phi ({x_i}) + b} \right) \ge 1 - {\xi _i},{\rm{\  }}{\xi _i} \ge 0,{\rm{\  }}i = 1,2,...,n.\]
The DWD method [11] uses a new criterion for the optimization problem [12].
We denote the functional margin as ${u_i} = {y_i}(w \cdot \phi ({x_i}) + b)$  and
let ${r_i} = {y_i}\left( {w \cdot \phi ({x_i}) + b} \right) + {\xi _i}$  be the adjusted distance of
the $i$-th data to the separating hyperplane. So the solution of DWD is given below:
\[\mathop {\min }\limits_{w,b,{\bf{\xi }}} \sum\limits_{i = 1}^n {\left( {\frac{1}{{{r_i}}} + C{\xi _i}} \right)} {\rm{\ \ \ }}s.t.{\rm{\ }}{r_i} = {y_i}\left( {w \cdot \phi ({x_i}) + b} \right) + {\xi _i},{\rm{\ }}{r_i} \ge 0,{\rm{\  }}{\xi _i} \ge 0,{\rm{\  }}{\left\| w \right\|^2} \le 1,{\rm{\  }}i = 1,2,...,n.\]
Furthermore, the LDM method optimizes the margin distribution to solve the optimization problem
[13][14]. Thus the optimization problem becomes
\begin{equation}
\begin{array}{l}
\mathop {\min }\limits_{w,{\bf{\xi }}} \frac{1}{2}{\left\| w \right\|^2} + \frac{{2{\lambda _1}}}{{{n^2}}}(n{w^T}X{X^T}w - {w^T}Xy{y^T}{X^T}w) - \frac{{{\lambda _2}}}{n}w \cdot (Xy) + C\sum\limits_{i = 1}^n {{\xi _i}} \\
s.t.{\rm{\ }}{y_i}w \cdot \phi ({x_i}) \ge 1 - {\xi _i},{\rm{\  }}{\xi _i} \ge 0,{\rm{\  }}i = 1,2,...,n,
\end{array}
\end{equation}
where $X = \left[ {\phi ({x_1}),\phi ({x_2}),...,\phi ({x_n})} \right]$, $y = {\left[ {{y_1},{y_2},...,{y_n}} \right]^T}$,
$y = \{  - 1, + 1\}$ is the label set. LDM offers two strategies to solve Formula (1):
for small and medium datasets, Formula (1) can be solved by the CD method [13], and for large-scale problem, ASGD is used [13][15][16].
ASGD solves the optimization objective by computing a noisy unbiased estimate of the gradient and it randomly
samples a subset of the training instances rather than all the data. Our method uses a similar implementation
of the CD and ASGD methods, which will be introduced in detail in the next section.
\section{The proposed e-DWSVR}
In this section, we describe the novel e-DWSVR method, which applies the idea of the mean of functional margin and
employs a similar solution as in LDM, that is, we use the CD strategy to handle general conditions and adopt the
ASGD strategy to deal with large-scale problems.
\subsection{Formulation for e-DWSVR }
For regression problems, the input is $X = \left[ {\phi ({x_1}),\phi ({x_2}),...,\phi ({x_n})} \right]$ and the
target is $y = {\left[ {{y_1},{y_2},...,{y_n}} \right]^T}$, where $y \in \{  - \infty , + \infty \}$. In order to
 simplify the complexity brought by the bias term, we enlarge the dimension of the vectors $w$  and $\phi ({x_i})$
 to handle the bias term [17], i.e., $w \leftarrow {[w,b]^T}$, $\phi ({x_i}) \leftarrow [\phi ({x_i}),{\bf{1}}]$.
 Thus the form of the regression function becomes $f(x) = w \cdot \phi (x)$. Then the margin in the regression analysis
 will be the distance of a point to the fitting hyperplane, i.e. $\left| {w \cdot \phi ({x_i}) - {y_i}} \right|/\left\| w \right\|.$
 Based on the concept of margin, we give the definition of functional margin.

\textbf{Definition 1:} functional margin in regression: $\gamma  = {\left( {w \cdot \phi ({x_i}) - {y_i}} \right)^2},{\rm{\ \ }}i = 1,...,n.$

The functional margin can describe the difference between the real values and the estimated value. It also has a significant
connection with the geometrical distance. In fact, the functional margin is the adjusted distance of the data to the
separating hyperplane. If the value of $w$ is determined, the ranking of all data points to the fitting surface with
respect to the margin can be decided by functional margin. Then the mean of the functional margin in regression is defined by Definition 2 as follows.

\textbf{Definition 2:} mean of the functional margin in regression: $\bar \gamma  = \frac{1}{n}\sum\limits_{i = 1}^n {{{\left( {w \cdot \phi ({x_i}) - {y_i}} \right)}^2}}.$

Based on the above definitions, we will not only maximize the minimum margin, but also minimize the mean of the
functional margin at the same time, in order to obtain a better tradeoff between the distribution of the whole data
and the distribution of support vectors. e-DWSVR considers the influence of all data to the fitting surface, as this
is closer to the actual distribution of the internal data.

To illustrate the difference of the optimal objectives between e-SVR and e-DWSVR, we use an example for comparison among linear regression, e-SVR and e-DWSVR on an artificial dataset. The distribution of the whole data is shown in Figure 1(a), where those sparse quadrate points are the noise and the red solid line with points is drawn by linear regression. Figure 1(b) is the enlarged version of the dotted box in Figure 1(a) for better observing the differences among the three methods. In Figure 1(b) the points show the distribution of most non-noisy data and the black dotted lines represent the data on the boundary. The red solid line with points is part of the line in Figure 1(a). The blue dashed curve is drawn by e-SVR and the red solid curve is drawn by e-DWSVR.

Obviously, the curve drawn by linear regression is far from the dataset, which indicates that the linear regression is more sensitive to noisy points, and this makes the curve in a wrong direction. e-SVR and e-DWSVR are not affected easily by noise points, so the blue dashed curve and the red solid curve are within area of non-noisy data. However, the regression model implemented by e-SVR is controlled by the support vectors (those boundary points). Once the internal data distribution is different from the edge points (which is the case in Figure 1(b)), the regression model will not achieve good performance. e-SVR produces the blue dashed curve which is different from the red solid curve. Because e-DWSVR considers the influence of all data to the fitting surface, it is obvious the red solid curve drawn by e-DWSVR is closer to the actual distribution of the internal data.

\begin{figure}[H]
  \setlength{\abovecaptionskip}{0cm}
  \setlength{\belowcaptionskip}{-0.4cm}
  \centering
  \setlength{\fboxrule}{0pt}
  \fbox{\rule[0cm]{0cm}{0cm}
  \includegraphics[height=5cm,width=12cm]{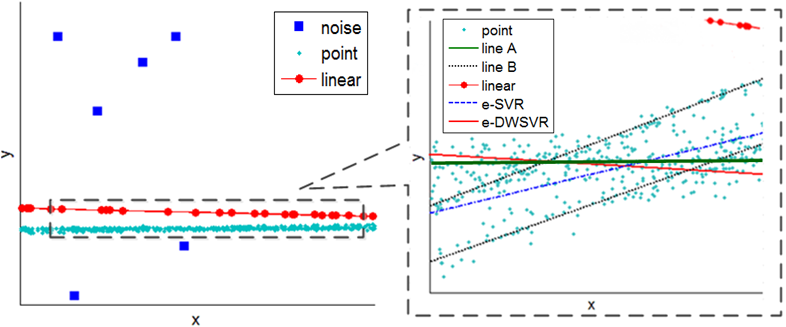}}\\
  \quad\quad\quad(a) The whole data with noise\quad\quad\quad (b) The distribution of data without noise\\
  \quad
  \caption{The fitting curves drawn by linear regression, e-SVR, and e-DWSVR. In Figure 1(b), 82.6\% of data points are evenly distributed across the line A, 16.5\% are evenly distributed on line B, where they are in parallel and have the different slope with the red solid curve. The other 0.8\% of data are outliers in Figure 1(a). Due to the outliers, the line produced by linear regression will be different from the correct curve (line A). The blue dashed curve produced by SVR will be very close to line B. While the red solid curve produced by eDWSVR is closer to line B, thus a better result can be produced.}
\end{figure}

As in the soft-margin e-SVR [1], we also consider complicated conditions. So, the final optimization function has the following form:
\[\begin{array}{l}
\mathop {\min }\limits_{w,\xi ,{\xi ^*}} \frac{1}{2}{\left\| w \right\|^2} + {\lambda _1}\bar \gamma  + C\sum\limits_{i = 1}^n {\left( {{\xi _i} + \xi _i^*} \right)} \\
s.t.{\rm{\ }}{y_i} - w \cdot \phi ({x_i}) \le \varepsilon  + {\xi _i}{\rm{,\ \ }}w \cdot \phi ({x_i}) + {y_i} \le \varepsilon  + \xi _i^*{\rm{,\ \ }}{\xi _i},\xi _i^* \ge 0,{\rm{\ \ }}i = 1,2,...,n.
\end{array}\]

\subsection{e-DWSVR for medium-scale regression with kernel.}
Considering the mean of the functional margin in the constrained minimization problem, we can obtain the following form:
\begin{equation}
\begin{array}{l}
\mathop {\min }\limits_{w,{\bf{\xi }},{{\bf{\xi }}^*}} \frac{1}{2}{\left\| w \right\|^2} + \frac{{{\lambda _1}}}{n}\left( {{w^T}X{X^T}w - 2{{\left( {Xy} \right)}^T}w} \right) + C\sum\limits_{i = 1}^n {\left( {{\xi _i} + \xi _i^*} \right)} \\
s.t.{\rm{\ }}{y_i} - w \cdot \phi ({x_i}) \le \varepsilon  + {\xi _i},{\rm{\ \ }}w \cdot \phi ({x_i}) + {y_i} \le \varepsilon  + \xi _i^*,{\rm{\ \ }}{\xi _i},\xi _i^* \ge 0,{\rm{\ \ }}i = 1,2,...,n.
\end{array}
\end{equation}
Here we omit the $y{y^T}$ term in $\bar \gamma$  because it is regarded as a constant in an optimization problem.
Obviously, the high dimensionality of $\phi$ and  its complicated form makes Formula (2) more intractable.
To simplify this formula, we take the suggestion from [18] and the optimal solution $w$ in
LDM [13]. We first give the following theorem.

\textbf{Theorem 1.} The optimal solution $w$ for Formula (2) can be represented as the following form:
where $\alpha  = {\left[ {{\alpha _1},{\alpha _2},...,{\alpha _n}} \right]^T}$ and
${\alpha ^{\rm{*}}} = {\left[ {{\alpha _1}^{\rm{*}},{\alpha _2}^{\rm{*}},...,{\alpha _n}^{\rm{*}}} \right]^T}$ are the parameters of e-DWSVR.

\textbf{PROOF.} We assume that $w$ can be divided into the span of $\phi ({x_i})$ and an orthogonal vector, i.e.,
\[w = \sum\limits_{i = 1}^n {\left( {{\alpha _i} - \alpha _i^*} \right)}  \cdot \phi ({x_i}) + u = X\left( {\alpha  - {\alpha ^*}} \right) + u, \]
where $u$ satisfies $\phi ({x_k}) \cdot u = 0$ for all $u$, that is, $X \cdot u = 0.$ Then we obtain the following formula:
Hence, the second term of Formula (2) is independent of $u$ considering Formula (4). Furthermore, note that the
constraints of Formula (2) are also independent of $u$, so the last term of Formula (2) can be considered as
independent of $u$. To simplify the first term, we have
In the above "$\ge $" becomes "${\rm{ = }}$" if and only if $u = 0$. Thus, setting $u = 0$ strictly reduces the first
term of Formula (2) without affecting the rest terms. Based on all above, $w$ in Formula (2) can be represented
in the form of Formula (3). \textbf{Q.E.D.}

According to Theorem 1, Formula (2) can be cast as
where $Q = 2{\lambda _1}{G^T}G/n + G$, $p =  - 2{\lambda _1}Gy/n$ and $G = {X^T}X.$ The minimization problem which
has been simplified can be transformed into a dual formulation with the Lagrange multipliers, so the Lagrange
function of Formula (5) leads to
where $\eta ,{\eta ^*},\beta ,{\beta ^*}$ are the Lagrange multipliers. To satisfy the KKT condition,
we set the partial derivatives $(\alpha ,{\alpha ^*},\xi ,{\xi ^*})$ to zero and obtain the following formulas:
By substituting Formulas (7) and (8) into Formula (6) and inspired by reference [19], Formula (5)
 can be written as follows to compute the values of $\left[ {\beta ,{\beta ^*}} \right]$ separately.
where $H = G{Q^{ - 1}}G$, and ${Q^{ - 1}}$ stands for the inverse matrix of $Q$ and $e$ means the all-one vector.

So Formula (9) can be solved by the CD method as in [20][21]. This method continuously
selects one variable for minimization and keeps others as constants. In our situation,
we minimize $\beta _k^{'} \in {\beta ^{'}}$ by fixing other $\beta _{l \ne k}^{'}$,
where ${\beta ^{'}} = \left[ {\beta ,{\beta ^*}} \right]$, and $\mathop {\min }\limits_t f({\beta ^{'}} + t{e_k}){\rm{\  \ }}s.t.{\rm{\ }}0 \le {\beta _k}^{'} + t \le C$
needs to be solved, where ${e_k} = {[\underbrace {0,...,0}_{k - 1},1,\underbrace {0,...,0}_{2n - k}]^T}$, and the
form of this sub-problem is
where ${h_{kk}}$ is the diagonal entry of $\left[ \begin{array}{l}
H{\rm{\ \ }} - H\\
 - H{\rm{\ \ \ }}H
\end{array} \right]$, and ${\left[ {\nabla f({\beta ^{'}})} \right]_k}$ is the $k$-th element of the gradient $\nabla f({\beta ^{'}})$. Obviously $f({\beta ^{'}})$ is independent of $t$, so we omit this part of Formula (10). Hence $f({\beta ^{'}} + t{e_k})$ is transformed into a simple quadratic function.

After calculating ${\beta ^{'}} = \left[ {\beta ,{\beta ^*}} \right]$, we can obtain $\left( {\alpha  - {\alpha ^*}} \right)$  according to Formula (7) as $\left( {{\alpha} - \alpha^*} \right) = {Q^{ - 1}}\left( {G\left( {\beta  - {\beta ^*}} \right) - p} \right) = {Q^{ - 1}}G\left( {\frac{{{\lambda _2}}}{n}Y + \left( {\beta  - {\beta ^*}} \right)} \right).$
Therefore, we obtain the final fitting function as $f(x) = \sum\limits_{i = 1}^n {({\alpha _i} - \alpha _i^*)} K({x_i},x).$

Algorithm 1 presents the detailed steps of the CD method used for efficiently updating ${\beta ^{'}} = \left[ {\beta ,{\beta ^*}} \right]$.
\begin{algorithm}
  \footnotesize
  \caption{\footnotesize Kernel e-DWSVR}
  \textbf{Input:} Dataset $X$, $\lambda_1$, $C$, $\varepsilon$\\
  \textbf{Output:} $\alpha  - {\alpha ^*}$\\
  \textbf{Initialization:} ${\beta ^{'}} = {\bf{0}},{\rm{\ }}\left( {\alpha  - {\alpha ^*}} \right) = \frac{{2{\lambda _1}}}{n}{Q^{ - 1}}Gy,{\rm{\ }}A = {Q^{ - 1}}G,{\rm{\ }}{h_{kk}} = e_k^TG{Q^{ - 1}}G{e_k};$
  \begin{algorithmic}[1]
  \For { $inter = 1,2,...,maxInter$ }
      \For { $k = 1,2,...,2n$ }
          \State ${\left[ {\nabla f({\beta ^{'}})} \right]_k} \leftarrow \varepsilon  + \left( {G\left( {\alpha  - {\alpha ^*}} \right) - {y_k}} \right);$ \quad\quad\quad if\quad $k = 1,2,...,n$
          \State ${\left[ {\nabla f({\beta ^{'}})} \right]_k} \leftarrow \varepsilon  - \left( {G\left( {\alpha  - {\alpha ^*}} \right) - {y_{k - n}}} \right);$ \quad\quad if\quad $k = n+1,n+2,...,2n$
          \State $\beta _k^{'old} \leftarrow \beta _k^{'};$
          \State $\beta _k^{'} \leftarrow \min (\max (\beta _k^{'} - \frac{{{{\left[ {\nabla f({\beta ^{'}})} \right]}_k}}}{{{h_{kk}}}},0),C);$
          \For { $i = 1,2,...,n$ }
              \State $\left( {{\alpha _i} - \alpha _i^*} \right) \leftarrow \left( {{\alpha _i} - \alpha _i^*} \right) + \left( {\beta _k^{'} - \beta _k^{'old}} \right)A{e_k};$\quad\quad if\quad $k = 1,2,...,n$
              \State $\left( {{\alpha _i} - \alpha _i^*} \right) \leftarrow \left( {{\alpha _i} - \alpha _i^*} \right) - \left( {\beta _k^{'} - \beta _k^{'old}} \right)A{e_k};$ \quad\quad if\quad $k = n+1,n+2,...,2n$
          \EndFor
      \EndFor
      \If {${\beta ^{'}}$ converge} break; \EndIf
  \EndFor
  \end{algorithmic}
\end{algorithm}

\subsection{e-DWSVR for large-scale regression.}
Although the CD method can solve e-DWSVR efficiently, it is not a best strategy for dealing with large-scale problems.
To further improve the scalability of e-DWSVR, we also apply ASGD method to e-DWSVR.

We reformulate Formula (2) as follows to solve large-scale problems,
\begin{align*}
\begin{array}{l}
\mathop {\min }\limits_w g(w) = \frac{1}{2}{\left\| w \right\|^2} + \frac{{{\lambda _1}}}{n}\left( {{w^T}{X^T}Xw - 2{{\left( {Xy} \right)}^T}w} \right)\\
 + C\sum\limits_{i = 1}^n {\max \left\{ {0,{y_i} - w \cdot {x_i} - \varepsilon ,w \cdot {x_i} - {y_i} - \varepsilon } \right\}}.\tag{11}
\end{array}
\end{align*}
Computing the gradient of $w$ in Formula (11) is time consuming because we need all the training instances for
computation. Considering this issue, we use Stochastic Gradient Descent (SGD) [22] to compute a noisy
unbiased estimation of the gradient, and this is done by randomly sampling part of the training instances.

Therefore, we give an unbiased estimation of the gradient $\nabla g(w)$ in our case. For denoting the last term of (11)
formally, we define a function $s(w)$ that has different values under different constraint conditions, as shown below:
\[s(w) = \left\{ \begin{array}{l}
 - {x_i},{\rm{\quad\quad\ }}i \in {I_1}\\
{x_i},{\rm{\quad\quad\quad}}i \in {I_2}\\
0,{\rm{\quad\quad\quad}}otherwise
\end{array} \right.{\rm{\ \  }},i = 1,2,...,n,\]
where ${I_1} \equiv \left\{ {i\left| {{y_i} - w \cdot {x_i} \le \varepsilon } \right.} \right\}{\rm{,\ }}{I_2} \equiv \left\{ {i\left| {w \cdot {x_i} - {y_i} \le \varepsilon } \right.} \right\}$.

\textbf{Theorem 2.} An unbiased estimation of the gradient $\nabla g(w)$ has the following form:
\[\nabla g(w,{x_i}) = 2{\lambda _1}{x_i}x_i^Tw + w - 2{\lambda _1}{y_i}{x_i} + nC \cdot s(w),\]
where one instance $({x_i},{y_i})$ is sampled randomly from the training set.

\textbf{PROOF:} The gradient of $g(w)$ has the form $\nabla g(w) = Qw + p + C\sum\limits_{i = 1}^n {s(w)}$, where $Q = 2{\lambda _1}{X^T}X/n + I$ and ${\rm{ }}p =  - 2{\lambda _1}Xy/n$. Note that
Considering the linearity of expectation, and with Formula (12), we have
This leads to a conclusion that $\nabla g(w,{x_i})$ is a noisy unbiased estimation of the gradient. \textbf{Q.E.D.}

Then, the stochastic gradient can be updated iteratively with the following form:
where ${\varphi _t}$ is the learning rate at the $t$-th iteration.

The good choice for ${\varphi _t}$  can be obtained by the form ${\varphi _t} = {\varphi _0}{(1 + a{\varphi _0}t)^{ - c}}$ by ASGD,
where $a$, ${\varphi _0}$, and $c$ are constants [23]. And we compute ${\bar w_t}$ at each iteration in
addition to updating the ordinary stochastic gradient in Formula (13) as
${\bar w_t}{\rm{ = }}\frac{1}{{t - {t_0}}}\sum\limits_{i = {t_0} + 1}^t {{w_t}}$, where ${t_0}$ decides when to perform
the averaging process. This average can also be calculated in a recursive manner as ${\bar w_{t + 1}} = {\delta _t}{w_{t + 1}} + (1 - {\delta _t}){\bar w_t}$,
where ${\delta _t}$ is the rate of averaging and ${\delta _t} = 1/\max \{ 1,t - {t_0}\}.$

Finally, Algorithm 2 provides the detailed steps of ASGD for large-scale problems, where $T*n$ determines the number of iteration. $T$ means the adjustment of the whole times of iteration which is set as same as [13]; $u$ means $\nabla g(w,{x_i})$, and its initial value is 0.
\begin{algorithm}[H]
  \footnotesize
  \caption{\footnotesize Large-scale Kernel e-DWSVR}
  \textbf{Input:} Dataset $X$, $\lambda_1$, $\varepsilon $\\
  \textbf{Output:} $\bar w$\\
  \textbf{Initialization:} $u = 0,t = 1$
  \begin{algorithmic}[1]
  \While {$t \le T*n$}
  \State Randomly select one instance $({x_i},{y_i})$ from the training set;
  \State Compute $\nabla g(w,{x_i})$;
  \State ${w_{t + 1}} \leftarrow {\varphi _t}\nabla g(w,{x_i}) + {w_t}$;
  \State ${\bar w_{t + 1}} \leftarrow {\delta _t}{w_{t + 1}} + (1 - {\delta _t}){\bar w_t}$;
  \State $t +  + $;
  \EndWhile
  \end{algorithmic}
\end{algorithm}
\section{Experiments}
In this section, we report the effectiveness of e-DWSVR in comparison with other regression methods
to assess whether our method has better fitting quality.
\subsection{Experimental setup}
We select sixteen datasets from UCI [24] to evaluate these methods. This includes eight
medium-scale datasets and eight large-scale datasets.  The numbers of samples and features for each dataset
are indicated in Table 1 and 2. For instance, Slump (103 / 7) means the sample size is 103 and the number of
features is 7 for the Slump dataset. All features of datasets and target set are normalized into [0,1] for
balancing the influence of each feature. After normalization, we take a preprocess of the PCA with the
contribution to 95\% for feature extraction. During the construction of the regression model, we divide
the datasets into training sets and testing sets by 5-fold cross validation, and the experiments are
repeated for 30 times. For medium-scale datasets, we evaluate both the linear and RBF
kernels [1], and the average values and standard deviation are recorded. For large-scale
datasets, only the linear kernel is evaluated. In addition, we record the time cost.

e-SVR, NN [6], linear regression and logistic are compared with our method. For e-SVR and e-DWSVR,
parameters $C$ and $e$ are both needed when building models. So these parameters are optimized during
the experiments to select the best parameters for regression. For e-DWSVR, parameter $\lambda_1$ is fixed
to 1; parameter $T$, ${t_0}$, and ${\varphi _t}$ have the same values as in [13]. All optimization are processed on testing sets.

The mean square error (MSE) [25] and R-square (R2) [26] are commonly used evaluation metrics in regression, and thus they are chosen in our research for evaluation. All experiments are tested with MATLAB on a PC, which has a 2.50 GHz CPU and 8 GB memory.
\subsection{Result and discussion}
For medium-scale datasets, Figure 1 shows the results of MSE on all methods,
including linear kernel function and RBF kernel function for e-DWSVR and e-SVR. Table 1 summarizes the results of R2 on all methods.
In most datasets, e-DWSVR performs better than other methods. Obviously, the fitting quality of e-DWSVR is much better than e-SVR, which
means e-DWSVR is more competitive than e-SVR. In Table 2, the best R2 on each dataset is indicated in bold.
\begin{figure}[H]
  \setlength{\abovecaptionskip}{0cm}
  \setlength{\belowcaptionskip}{0cm}
  \centering
  \setlength{\fboxrule}{0pt}
  \fbox{\rule[0cm]{0cm}{0cm}
  \includegraphics[height=7.5cm,width=12cm]{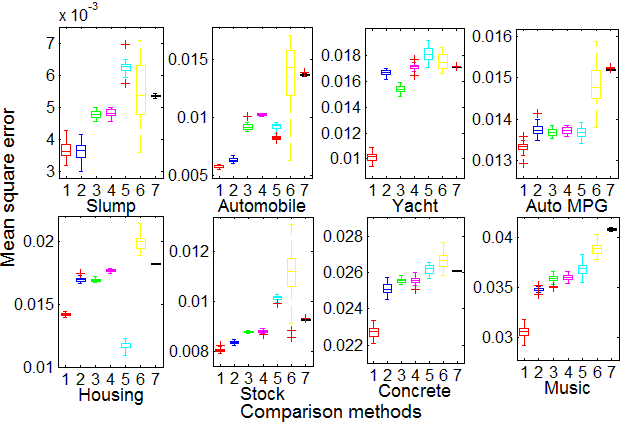}}\\
  \caption{The evaluation of MSE on the medium-scale datasets. It can be seen that e-DWSVR and e-SVR present the advanced results for RBF kernel. Also, e-DWSVR achieves the smallest MSE and largest R2 in most datasets,
    which means that e-DWSVR can provide better fitting quality. In addition, e-DWSVR performs significantly better than e-SVR for both linear kernel and RBF kernel. Values of x axis (from left to right) are comparison methods:
    1. e-DWSVR-RBF (red), 2. e-SVR-RBF (blue), 3. e-DWSVR-Linear (green), 4. e-SVR-Linear (purple), 5. linear regression (cyan), 6. NN (yellow), 7. logistic (black).}
\end{figure}
\begin{table}[H]
  \setlength{\abovecaptionskip}{0cm}
  \setlength{\belowcaptionskip}{0cm}
  \footnotesize
  \caption{The evaluation of R2 on the medium-scale datasets.}
  \label{sample-table}
  \centering
  \begin{tabular}{llllllll}
    \cmidrule{1-8}
    \tabincell{l}{Dataset\\(Samples / Features)}	&\tabincell{l}{eDWSVR\\(RBF)} &\tabincell{l}{eSVR\\(RBF)} &\tabincell{l}{eDWSVR\\(Linear)} &\tabincell{l}{eSVR\\(Linear)} &LINEAR &NN &Logistic\\
    \midrule
     Slump (103 / 7)	&\textbf{0.9539} &0.9224 &0.8739 &0.8641 &0.8067 &0.8220 &0.8336\\
    Automobile (205 / 26)	&\textbf{0.8865}	&0.8320	&0.7847	&0.7689	&0.7748	&0.7526 &0.7592\\
    Yacht (308 / 7)	&\textbf{0.8684}	&0.7526	&0.7397	&0.7083	&0.6574	&0.6990 &0.7007\\
    Auto MPG (398 / 8)	&\textbf{0.6900}	&0.6874	&0.6839	&0.6837	&0.6843	&0.6782 &0.6673\\
    Housing (506 / 14)	&0.6955	&0.6634	&0.6851	&0.6522	&\textbf{0.7048}	&0.6091 &0.6326\\
    Stock (536 / 9)	&\textbf{0.5687}	&0.5566	&0.5344	&0.5241	&0.4055	&0.3719 &0.4146\\
    Concrete (1030 / 9) &\textbf{0.5430}	&0.4590	&0.4300	&0.4289	&0.3810	&0.3583 &0.3710\\
    Music (1059 / 68)	&\textbf{0.3865}	&0.3628	&0.3084	&0.3074	&0.3331	&0.2485 &0.1472\\
    \bottomrule
  \end{tabular}
\end{table}
For large-scale datasets, Figure 2 and Table 2 summarize the results for the linear kernel. As one can see,
e-DWSVR performs better than other regression methods on some datasets, and it always performs better than
e-SVR. In addition, linear regression did not return the results on some datasets after 48 hours
(indicated as NA in Table 2).
\begin{figure}[H]
  \setlength{\abovecaptionskip}{0cm}
  \setlength{\belowcaptionskip}{-0.5cm}
  \centering
  \setlength{\fboxrule}{0pt}
  \fbox{\rule[0cm]{0cm}{0cm} \rule[0cm]{0cm}{0cm}
  \includegraphics[height=7.5cm,width=12cm]{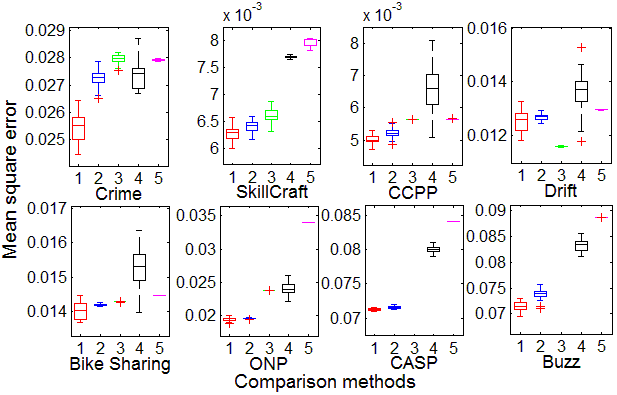}}\
  \caption{The evaluation of MSE on the large-scale datasets. Values of x axis (from left to right) are comparison methods: 1. e-DWSVR (red), 2. e-SVR (blue), 3. linear regression (green), 4. NN (black), 5. logistic (purple).}
\end{figure}
\begin{table}[H]
  \footnotesize
  \setlength{\abovecaptionskip}{0cm}
  \setlength{\belowcaptionskip}{0cm}
  \caption{The evaluation of R2 on the large-scale datasets.}
  \centering
  \begin{tabular}{llllll}
    \cmidrule{1-6}
    \tabincell{l}{Dataset\\(Samples / Features)}&e-DWSVR &e-SVR &LINEAR	&NN &Logistic\\
    \midrule
     Crime (1994 / 128)	&\textbf{0.5613}	&0.5348	&0.4430	&0.5244 &0.5350\\
     SkillCraft (3338 / 18)	&\textbf{0.7275} &0.7018 &0.7013 &0.6843 &0.6147\\
     CCPP (9568 / 4)	&\textbf{0.8634}	&0.8327	&0.8246	&0.7334 &0.8286\\
     Drift (13910 / 129)	&0.5620	&0.5636	&\textbf{0.5888}	&0.5490 &0.5574\\
     Bike sharing (17389 / 16)	&0.5867	&0.5825	&\textbf{0.5938}	&0.5395 &0.5465\\
     ONP (39797 / 61)	&\textbf{0.4598}	&0.4590	&0.3856	&0.3544 &0.2113\\
     CASP (45730 / 9)	&\textbf{0.2755}	&0.2608	&NA	&0.1836 &0.1657\\
     Buzz (140000 / 77)	&\textbf{0.2976}	&0.2591	&NA	&0.1529 &0.1373\\
    \bottomrule
  \end{tabular}
\end{table}
\subsection{Time cost and parameter effect}
In Figure 3 we present a comparison of CPU time taken between e-SVR and e-DWSVR on each dataset.
e-SVR for large-scale problems was implemented by the LIBLINEAR [17] package and e-DWSVR was implemented by ASGD. According to Figure 3, e-DWSVR cost less time than e-DWSVR on most datasets,
and it is only slightly slower than e-SVR on two datasets.

Furthermore, Figure 4 shows that the MSE on the medium-scale datasets does not have much difference with the change of the parameters.
This indicates that the performance of e-DWSVR is not sensitive to the parameters, which shows the robustness of e-DWSVR.
\begin{figure}[H]
  \setlength{\abovecaptionskip}{-0.2cm}
  \setlength{\belowcaptionskip}{-0.7cm}
  \centering
  \begin{minipage}{.48\linewidth}
  \setlength{\fboxrule}{0pt}
  \fbox{\rule[0cm]{0cm}{0cm} \rule[0cm]{0cm}{0cm}
  \includegraphics[height=2.5cm,width=5.5cm]{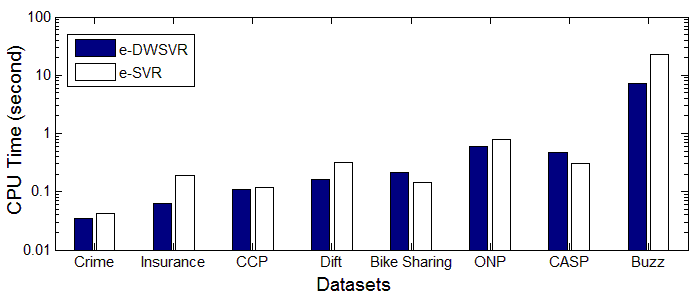}}\\
  \caption{CPU time on the large-scale datasets.}
  \end{minipage}
  \begin{minipage}{.48\linewidth}
  \setlength{\fboxrule}{0pt}
  \fbox{\rule[0cm]{0cm}{0cm} \rule[0cm]{0cm}{0cm}
  \includegraphics[height=2.5cm,width=6.5cm]{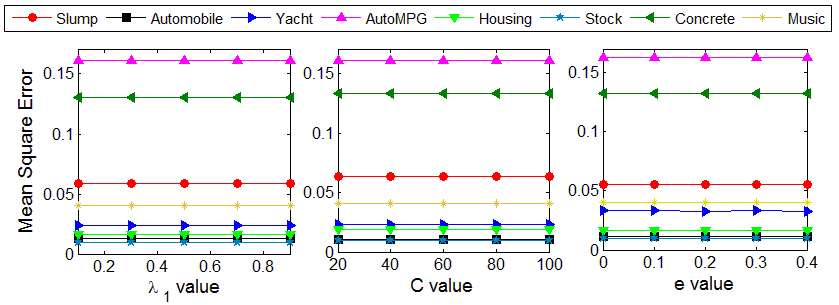}}\\
  \caption{Parameter effect on eDWSVR.}
  \end{minipage}
\end{figure}
\section{Conclusion and future work}
In this paper, we propose e-DWSVR, a novel and promising SVR algorithm. e-DWSVR overcomes the limitations of existing SVR systems and outperforms other regression methods on several benchmark datasets.
We envision great application potential of e-DWSVR in various problems, including feature extraction, anomaly detection, and complex data interpretation. In the near future, we will apply e-DWSVR to solve
several real-world problems, including bioinformatics anaysis and financial data prediction and anomaly detection.

\section{Acknowledgement}
We gratefully thank Dr Teng Zhang and Prof Zhi-Hua Zhou for providing the source code of “LDM” source code and their kind technical assistance. We also thank Prof Chih-Jen Lins team for providing the Libsvm and Liblinear packages and their support. This work is supported by the National Natural Science Foundation of China (Nos. 61472159, 61572227) and Development Project of Jilin Province of China (Nos.20140101180JC, 20160204022GX). This work is also partially supported by the 2015 Scottish Crucible Award funded by the Royal Society of Edinburgh and the 2016 PECE bursary provided by the Scottish Informatics \& Computer Science Alliance (SICSA).
\section*{References}
[1] V. Vapnik C. Cortes. Support vector machine. Machine Learning, 7(2002):1–28, 1995.

[2] J. D. Brown, M. F. Summers, and B. A. Johnson. Prediction of hydrogen and carbon chemical shifts from
rna using database mining and support vector regression. Journal of Biomolecular Nmr, 63(1):1–14, 2015.

[3] C. Y. Deng and H. W. Lin. Progressive and iterative approximation for least squares b-spline curve and
surface fitting. Computer-Aided Design, 47(1):32–44, 2014.

[4] M. V. Butz, G. K. M. Pedersen, and P. O. Stalph. Learning sensorimotor control structures with xcsf:
redundancy exploitation and dynamic control. In Proceedings of the 11th Annual Conference on Genetic
and Evolutionary Computation, pages 1171–1178, 2009.

[5] D. Liu, T.R. Li, and D. C. Liang. Incorporating logistic regression to decision-theoretic rough sets for
classifications. International Journal of Approximate Reasoning, 54(1):197–210, 2014.

[6] K. Hagiwara, T. Hayasaka, N. Toda, S. Usui, and K. Kuno. Upper bound of the expected training error of
neural network regression for a gaussian noise sequence. Neural Networks, 14(10):1419–1429, 2001.

[7] A. J. Smola and B. Scholkopf. A tutorial on support vector regression. Statistics and Computing,
14(3):199–222, 2004.

[8] B. Demir and L. Bruzzone. A multiple criteria active learning method for support vector regression. Pattern
Recognition, 47(7):2558–2567, 2014.

[9] B Scholkopf, A. J. Smola, R. C. Williamson, and P. L. Bartlett. New support vector algorithms. Neural
Computation, 12(5):1207–1245, 2000.

[10] B. Gu, V. S. Sheng, Z. J. Wang, D. Ho, S. Osman, and S. Li. Incremental learning for v-support vector
regression. Neural Networks the Official Journal of the International Neural Network Society, 67(C):140–
150, 2015.

[11] J. S. Marron. Distance-weighted discrimination. Journal of the American Statistical Association,
102(12):1267–1271, 2007.

[12] X. Y. Qiao and L. S. Zhang. Distance-weighted support vector machine. Statistics and Its Interface,
8(3):331–345, 2013.

[13] T. Zhang and Z. H. Zhou. Large margin distribution machine. In Proceedings of the 20th ACM SIGKDD
International Conference on Knowledge Discovery and Data Mining, pages 313–322, 2014.

[14] Z. H. Zhou. Large margin distribution learning. In Proceedings of the 6th Artificial Neural Networks in
Pattern Recognition, pages 1–11, 2014.

[15] B. T. Polyak and A. B. Juditsky. Acceleration of stochastic approximation by averaging. Siam Journal on
Control and Optimization, 30(4):838–855, 1992.

[16] T. Zhang. Solving large scale linear prediction problems using stochastic gradient descent algorithms. In
Proceedings of the 21st International Conference on Machine Learning, pages 919–926, 2004.

[17] R. E. Fan, K. W. Chang, C. J. Hsieh, X. R. Wang, and C. J. Lin. Liblinear: A library for large linear
classification. Journal of Machine Learning Research, 9(12):1871–1874, 2010.

[18] B. Scholkopf and A. Smola. Learning with kernels : support vector machines, regularization, optimization,
and beyond. MIT Press, 2002.

[19] C. C. Chang and C. J. Lin. Libsvm: A library for support vector machines. Acm Transactions on Intelligent
Systems and Technology, 2(3):389–396, 2011.

[20] G. X. Yuan, C. H. Ho, and C. J. Lin. Recent advances of large-scale linear classification. Proceedings of
the IEEE, 100(9):2584–2603, 2012.

[21] Z. Q. Luo and P. Tseng. On the convergence of the coordinate descent method for convex differentiable
minimization. Journal of Optimization Theory and Applications, 72(1):7–35, 1991.

[22] L. Bottou. Large-Scale Machine Learning with Stochastic Gradient Descent. Physica-Verlag HD, 2010.

[23] W. Xu. Towards optimal one pass large scale learning with averaged stochastic gradient descent. Computer
Science, 2011.

[24] M. Lichman. UCI machine learning repository. http://archive.ics.uci.edu/ml, 2013.

[25] D. N. Guo, S. Shamai, and S. Verdu. Mutual information and minimum mean-square error in gaussian
channels. IEEE Transactions on Information Theory, 51(4):1261–1282, 2005.

[26] W. Wen, R. Luo, X. J. Tang, L. Tang, H. X. Huang, X. Y. Wen, S. Hu, and B. Peng. Age-related
progression of arterial stiffness and its elevated positive association with blood pressure in healthy people.
Atherosclerosis, 238(1):147–152, 2015.
\end{document}